\documentclass[11pt]{article}

\usepackage[preprint]{acl}

\usepackage{times}
\usepackage{latexsym}

\usepackage[T1]{fontenc}

\usepackage[utf8]{inputenc}

\usepackage{microtype}

\usepackage{inconsolata}

\usepackage{graphicx}
\usepackage{booktabs}
\usepackage{amsmath}

\title{Efficient Multilingual Dialogue Processing via Translation Pipelines and Distilled Language Models}

\author{\textnormal{Santiago Martínez Novoa, Nicolás Rozo Fajardo,} \\
        \textnormal{Diego Alejandro González Vargas, Nicolás Bedoya Figueroa} \\
        \\
        Universidad de los Andes, Bogotá, Colombia \\
        \texttt{\{s.martinezn, n.rozo, da.gonzalezv1, n.bedoyaf\}@uniandes.edu.co}}

\begin{document}
\maketitle

\begin{abstract}
This paper presents team Kl33n3x's multilingual dialogue summarization and question answering system developed for the NLPAI4Health 2025 shared task. The approach employs a three-stage pipeline: forward translation from Indic languages to English, multitask text generation using a 2.55B parameter distilled language model, and reverse translation back to source languages. By leveraging knowledge distillation techniques, this work demonstrates that compact models can achieve highly competitive performance across nine languages. The system achieved strong win rates across the competition's tasks, with particularly robust performance on Marathi (86.7\% QnA), Tamil (86.7\% QnA), and Hindi (80.0\% QnA), demonstrating the effectiveness of translation-based approaches for low-resource language processing without task-specific fine-tuning.
\end{abstract}

\section{Introduction}

Multilingual dialogue understanding presents significant challenges, particularly for low-resource Indic languages \cite{joshi-etal-2020-state, kakwani-etal-2020-indicnlpsuite}. While recent advances in large language models (LLMs) have demonstrated impressive capabilities in English, their performance on other languages remains limited \cite{lauscher-etal-2020-zero,huang2025surveylargelanguagemodels}. These limitations are particularly pronounced when dealing with conversational data, which requires understanding context, speaker turns, and discourse structure \cite{jawale2024humanconversationsspeciallarge, zhang2020dialogptlargescalegenerativepretraining}.

This paper describes the system developed by team Kl33n3x for the NLPAI4Health 2025 shared task\footnote{\url{https://nlpai4health.com/}}, which focuses on generating summaries and answering questions from multilingual dialogues across eight Indic languages (Marathi, Kannada, Gujarati, Telugu, Tamil, Bangla, Hindi, and Assamese) plus English. The key insight is that leveraging high-quality translation models combined with powerful English-centric generative models can outperform direct multilingual approaches, especially when training data quality is questionable.

\section{System Architecture}

For the first implementation, the idea was to finetune an LLM in order to acquire a model that was trained for perfoming the tasks required. For this, two options were considered: a complete finetune of the model or a low-rank adaptation (LoRA) \citep{hu2021loralowrankadaptationlarge}. The first option was discarded for two main reasons: computational cost and training time, both aspects of this solution were really expensive, therefore, it was decided to continue with LoRA finetune. This solution seemed to be promising as it reduced training time. However, evaluation revealed that the LoRA-adapted model struggled with maintaining factual consistency across long dialogues and exhibited lower performance on specialized medical terminology, falling short of the required clinical accuracy standards. Because of this, it was necessary to pivot to another approach.

Thus, the final implemented system comprises three main components that form a sequential pipeline. Each component is carefully chosen to maximize both quality and computational efficiency.

\subsection{Forward Translation Model}

For translating dialogues from Indic languages to English, the system employs \textit{prajdabre/rotary-indictrans2-indic-en-dist-200M}, based on the IndicTrans2 architecture. This model extends foundational sequence-to-sequence architectures like MarianMT and mBART50, optimized specifically for Indic-English language pairs.

The model employs rotary position embeddings (RoPE) \citep{su2021roformer}, which improve generalization across sequence lengths and preserve positional information in continuous space. The architecture uses standard transformer self-attention and cross-attention mechanisms to encode source text into contextual embeddings and decode them into fluent English sentences.

The decision to use a model to translate first is rooted in performance considerations. While multilingual summarization and question answering models exist, they may exhibit lower robustness and consistency across the different Indic languages required. In contrast, English summarization and QA models benefit from significantly larger training corpus, stronger evaluation benchmarks, and more advanced domain adaptation techniques. By first translating all dialogues into English using a highly specialized and accurate Indic to English translation model, the system can leverage English-language models for summarization and question answering. Empirically and conceptually, this pipeline yields higher overall quality than relying on a single multilingual model with a wider language coverage but weaker performance in each individual language, thereby prioritizing output accuracy and reliability in a clinical context.

This translate-first approach is consistent with findings in cross-lingual NLP showing that specialized translation models followed by monolingual task-specific models often outperform end-to-end multilingual systems, particularly for low-resource languages \citep{hu2020xtrememassivelymultilingualmultitask,conneau2020unsupervised}.

\subsection{Multitask Generation Model}

At the core of the system is a distilled version of Qwen3-4B-Instruct. Specifically, the implementation utilizes \textit{unsloth/Qwen3-4B-Instruct-2507-unsloth-bnb-4bit}, which employs 4-bit quantization and optimization techniques from the Unsloth framework. The actual number of parameters in this distilled model is 2.55B, achieved through knowledge distillation techniques \citep{hinton2015distilling, sanh2019distilbert} that compress the original 4B parameter model while retaining most of its capabilities.

Knowledge distillation works by training a smaller "student" model to mimic the behavior of a larger "teacher" model, typically by matching output distributions or intermediate representations \citep{gou2021knowledge}. This approach allows the system to maintain strong performance while significantly reducing computational requirements and memory footprint,critical considerations for practical deployment.

This single model is employed for all three tasks (narrative summarization, structured summarization, and question answering) through task-specific prompting.

A decisive factor in the selection of this model is its exceptionally large context window, which supports sequences of up to approximately 256k tokens  \citep{yang2025qwen3technicalreport}. This capability is particularly critical for medical  dialogue processing, as previous work has shown that models with limited context windows cannot capture entire clinical conversations \citep{FraileNavarro2025}. 
Models with limited context windows would require truncation or sliding-window strategies, both of which risk losing clinically relevant information and may  result in poor utilization of information in the middle of long contexts  \citep{liu2023lostmiddlelanguagemodels}. Such dialogues could become even longer after translation  into English due to tokenization effects. By contrast, the large-context capacity of the Qwen architecture allows the entire conversational history to be processed in a single forward pass, enabling the model to ground its summaries and answers in the full dialogue context and to preserve discourse-level coherence and factual consistency.

Finally, the use of a single pre-trained, instruction-tuned model across all tasks offers significant practical advantages over maintaining separate specialized models. The selected 4-bit quantized version reduces GPU memory requirements to approximately 6GB compared to 18-24GB for loading multiple model instances simultaneously. This approach simplifies deployment architecture and maintenance—critical considerations for resource-constrained clinical settings. While specialized models fine-tuned for individual tasks might achieve higher performance on specific benchmarks, the instruction-following capabilities of Qwen3-4B-Instruct enable it to handle diverse tasks through prompt engineering alone \citep{wei2022finetunedlanguagemodelszeroshot, ouyang2022traininglanguagemodelsfollow}, making the computational and operational trade-offs favorable for our deployment scenario.

\subsection{Reverse Translation Model}

For translating generated English outputs back to source languages, the system uses \textit{prajdabre/rotary-indictrans2-en-indic-dist-200M}. This model mirrors the forward translation architecture but operates in the opposite direction.

\subsection{Preprocessing Pipeline}

The preprocessing pipeline begins with dialogue formatting, where raw dialogue data is structured with clear speaker attribution and turn boundaries. For both translation models, special language tokens are added to indicate source and target languages following IndicTrans2 conventions (e.g., \texttt{<2en>} for translation to English).

Text normalization includes standard procedures such as removing excessive whitespace, handling special characters, and ensuring consistent Unicode encoding. The pipeline preserves dialogue structure markers (speaker names, turn indicators) throughout the translation process to maintain conversational coherence.

For the generation model, task-specific instructions are prepended to dialogues in a standardized format, informing the model whether to produce narrative summaries, structured summaries, or question-answer pairs. Maximum sequence length constraints are enforced at each stage.

The reverse translation stage applies the appropriate language code for the target Indic language and handles any formatting artifacts introduced during generation, such as removing English discourse markers or adjusting punctuation conventions for the target language.

Each stage operates independently, allowing for potential optimization or replacement of individual components without affecting the overall architecture.

\subsection{Implementation Details}

The system was implemented using PyTorch and the HuggingFace Transformers 
library. The distilled translation models were loaded directly without 
additional quantization. The generation model was loaded in its 4-bit 
quantized version as provided in its pretrained checkpoint, reducing memory 
requirements to approximately 6GB.

Inference was performed with greedy decoding for both translation and 
generation steps to ensure deterministic outputs and maximize reproducibility. 
For the translation models, input sequences were truncated to 2048 tokens 
when necessary, with output generation limited to 2048 tokens. The generation 
model was configured to produce up to 3000 tokens per response.

Batch processing was employed where possible to improve throughput, with 
batch sizes adjusted based on available GPU memory. The generation model's 
large context window (256k tokens) was sufficient to process most dialogues 
in the dataset without requiring truncation or sliding window approaches 
for the summarization and question answering tasks. However, extremely long 
dialogues were truncated during the initial translation step to fit within 
the translation models' 2048-token input limit.

\section{Results and Analysis}

\subsection{Competition Performance}

The system demonstrated strong performance across the NLPAI4Health shared task, achieving competitive win rates across multiple languages and tasks. Table \ref{tab:competition_results} shows the system's win rates in head-to-head comparisons with other participating systems.

\begin{table*}[t]
\centering
\small
\begin{tabular}{lccc}
\toprule
\textbf{Language} & \textbf{QnA} & \textbf{Summary (Text)} & \textbf{Summary (KnV)} \\
\midrule
Marathi & 86.7\% & 60.0\% & 60.0\% \\
Tamil & 86.7\% & 60.0\% & 60.0\% \\
Hindi & 80.0\% & 66.7\% & 53.3\% \\
Assamese & 73.3\% & 40.0\% & 60.0\% \\
Bangla & 66.7\% & 40.0\% & 40.0\% \\
Gujarati & 60.0\% & 73.3\% & 53.3\% \\
Kannada & 60.0\% & 66.7\% & 53.3\% \\
English & 46.7\% & 46.7\% & 53.3\% \\
Telugu & 13.3\% & 53.3\% & 66.7\% \\
\bottomrule
\end{tabular}
\caption{Competition win rates showing the system's performance across tasks. The approach achieved particularly strong results on Marathi and Tamil for question answering, while showing more balanced performance across all tasks for other languages.}
\label{tab:competition_results}
\end{table*}

The results reveal several interesting patterns. The system excelled at question answering for languages with strong translation model support (Marathi, Tamil, Hindi), suggesting that translation quality is a key factor in downstream task performance. For Telugu, while QnA performance was lower, the system performed well on summarization tasks, indicating that different task types have varying sensitivities to translation artifacts. The structured summarization (KnV) task showed more consistent performance across languages, possibly because the key-value format is more robust to minor translation errors.

\subsection{Automatic Evaluation Metrics}

Table \ref{tab:auto_metrics} presents detailed automatic evaluation scores across tasks and languages. The system achieved F1 scores ranging from 0.43 to 0.67 for question answering, 0.81 to 0.92 for narrative summarization, and 0.35 to 0.43 for structured summarization.

\begin{table*}[t]
\centering
\footnotesize
\begin{tabular}{lcccccc}
\toprule
\textbf{Language} & \multicolumn{2}{c}{\textbf{QnA}} & \multicolumn{2}{c}{\textbf{Summary (Text)}} & \multicolumn{2}{c}{\textbf{Summary (KnV)}} \\
& \textbf{F1} & \textbf{BERT} & \textbf{F1} & \textbf{BERT} & \textbf{F1} & \textbf{BERT} \\
\midrule
English & 0.668 & 0.850 & 0.900 & 0.839 & 0.427 & 0.908 \\
Hindi & 0.617 & 0.866 & 0.922 & 0.838 & 0.400 & 0.904 \\
Marathi & 0.598 & 0.850 & 0.815 & 0.811 & 0.390 & 0.911 \\
Tamil & 0.564 & 0.846 & 0.911 & 0.839 & 0.379 & 0.914 \\
Telugu & 0.565 & 0.851 & 0.893 & 0.834 & 0.364 & 0.911 \\
Kannada & 0.558 & 0.849 & 0.879 & 0.828 & 0.372 & 0.915 \\
Gujarati & 0.529 & 0.849 & 0.895 & 0.835 & 0.359 & 0.913 \\
Bangla & 0.480 & 0.829 & 0.852 & 0.818 & 0.346 & 0.903 \\
Assamese & 0.428 & 0.824 & 0.879 & 0.830 & 0.383 & 0.915 \\
\bottomrule
\end{tabular}
\caption{Automatic evaluation metrics (F1 and BERTScore) across tasks. The system achieves particularly strong F1 scores on narrative summarization (0.81--0.92), demonstrating effective translation and generation capabilities.}
\label{tab:auto_metrics}
\end{table*}

The high BERTScore values (0.82--0.92) across all tasks indicate strong semantic similarity between generated and reference outputs, even when exact lexical overlap (as measured by F1) is lower. This suggests the system captures the meaning of dialogues effectively, even if surface realization differs from references.

\subsection{Cross-Task Performance Patterns}

Comparing performance across tasks reveals that narrative summarization benefits most from the translation-based approach, likely because summarization is more robust to minor translation artifacts than extractive question answering. The gap between F1 scores for narrative (0.81--0.92) versus structured (0.35--0.43) summarization highlights the additional challenge of format conversion and information extraction required for key-value outputs.

Languages with extensive IndicTrans2 training data (Hindi, Bengali) show consistently strong performance across tasks, reinforcing the importance of high-quality translation as the foundation of the pipeline.

\section{Error Analysis}

Several categories of errors were identified in the system outputs. Translation 
artifacts occasionally appear when domain-specific terms or cultural references 
are mistranslated, affecting downstream task performance. This was particularly 
evident in Telugu, where the system struggled more with question answering 
despite reasonable summarization performance.

For format consistency, the structured summarization task sometimes produces 
outputs that deviate from expected key-value formats, particularly for complex 
dialogues. Translation input truncation occasionally affects extremely long 
dialogues, as the translation models' 2048-token input limit may result in 
truncation of very lengthy conversations before they reach the generation stage. 
While the generation model's 256k-token context window is sufficient for all 
dialogues in the dataset, this upstream truncation can lead to information loss 
for edge cases. Finally, dialogue structure features such as speaker attribution 
and turn-taking patterns are sometimes lost in translation, affecting coherence 
in generated summaries.

\section{Limitations}

While the system demonstrates strong performance, several limitations should be acknowledged. The approach is fundamentally limited by the quality of the translation models, as errors in translation propagate through the pipeline and cannot be fully corrected downstream. This is evident in the lower performance on Telugu question answering.

Despite using a distilled 2.55B parameter model, the three-stage pipeline (forward translation, generation, reverse translation) requires more computation and introduces additional latency compared to a direct multilingual approach, if such an approach were sufficiently accurate for these low-resource languages.

Translation may lose cultural context, idioms, or language-specific discourse markers that could be important for understanding dialogue content. The sequential pipeline introduces latency that may be problematic for real-time applications, though this could be mitigated through batching and optimization.

Finally, while avoiding fine-tuning on noisy data was beneficial in this case, future work with high-quality annotated data might benefit from task-specific fine-tuning of the generation component.

\section{Conclusion}

This paper presented an efficient multilingual dialogue summarization and question answering system for the NLPAI4Health shared task. The key findings demonstrate that a distilled 2.55B parameter model can achieve competitive performance with larger models while offering significant computational advantages. Translation-based approaches remain viable for multilingual NLP tasks, particularly when direct multilingual models are unavailable or training data quality is questionable.

The system achieved strong competition performance with win rates of 86.7\% (Marathi, Tamil) and 80\% (Hindi) for question answering, alongside F1 scores of 0.81--0.92 on narrative summarization tasks. These results demonstrate the effectiveness of combining high-quality translation with powerful distilled generative models. Future work will explore direct multilingual approaches as improved models and datasets become available, as well as methods to reduce translation dependency and improve efficiency further.

\section*{Acknowledgements}

The authors thank the organizers of the NLPAI4Health 2025 shared task for providing the datasets and evaluation infrastructure. The authors also acknowledge Universidad de los Andes for computational resources.

\bibliography{custom}

\end{document}